# Performance Evaluation of Statistical Approaches for Text-Independent Speaker Recognition Using Source Feature

[1]R. Rajeswara Rao, [2]V. Kamakshi Prasad, [3]A. Nagesh
[1]DVR College of Engineering & Technology, Department of CSE, Hyderabad, AP, India
raob4u@yahoo.com
[2]JNT University, Hyderabad, AP, India. [3]MG Institute of Technology, Hyderabad, AP, India
akknagesh@rediffmail.com

**Abstract –** This paper introduces the performance evaluation of statistical approaches for Text-Independent speaker recognition system using source feature. Linear prediction (LP) residual is used as a representation of excitation information in speech. The speaker-specific information in the excitation of voiced speech is captured using statistical approaches such as Gaussian Mixture Models (GMMs) and Hidden Markov Models (HMMs). The decrease in the error during training and recognizing speakers during testing phase close to 100% accuracy demonstrates that the excitation component of speech contains speaker-specific information and is indeed being effectively captured by continuous Ergodic HMM than GMM. The performance of the speaker recognition system is evaluated on GMM and 2-state ergodic HMM with different mixture components and test speech duration. We demonstrate the speaker recognition studies on TIMIT database for both GMM and Ergodic HMM.

**Index Terms:** *Ergodic, LP residual, MFCC, Speaker.*

## 1. INTRODUCTION

Within the past decade, technological advances such as telebanking and remote collaborative data processing over large computer networks have increased the demand for improved methods of information security. For personal information including medical records, bank accounts and credit history, the ability to verify the identity of individuals attempting to access such data is critical. To date, low-cost methods such as passwords, personal identification numbers and magnetic cards have been widely used. More advanced security measures have also been developed (e.g., face recognizers, retinal scanners, as well as automatic finger print analyzers). The uses of these procedures have been limited by both cost and ease of use. In recent years, speaker recognition (recognizing a person from his/her voice by a machine) and verification algorithms have received considerable attention. There are several reasons for this interest. In particular, speech provides a convenient and natural form of input, conveys a significant amount of speaker dependent information.

Speech is a composite signal which carries information about the message, the speaker identity and the language identity [1], [2]. It is difficult to isolate the speaker specific features alone from the signal. The speaker characteristics present in the signal can be attributed to the anatomical and the behavioural aspects of the speech production mechanism. The representation of the behavioural characteristics is a difficult task, and usually requires large amount of data.

Automatic speaker recognition systems rely mainly on features derived from the physiological characteristics of the speaker.

Speech is produced as sequence of sounds. Hence the state of the vocal folds, shape and size of various articulators, change over time to reflect the sound being produced. To produce a particular sound the articulators have to be positioned in a particular way. When different speakers try to produce same sound, though their vocal tracts are positioned in a similar manner, the actual vocal tract shapes will be different due to differences in the anatomical structure of the vocal tract. System features represent the structure of vocal tract. The movements of vocal folds vary from one speaker to another, the manner and speed in which the vocal folds close also varies across speakers. As a result different voices are produced by different speakers. The variations in the vibrations of the vocal folds represent the source features.

The theory of Linear Prediction (LP) is closely linked to modelling of the vocal tract system, and relies upon the fact that a particular speech sample may be predicted by a linear combination of previous samples. The number of previous samples used for prediction is known as the order of the prediction. The weights applied to each of the previous speech samples are known as Linear Prediction Coefficients (LPC). They are calculated so as to minimize the prediction error [4].

A study into the use of LPC for speaker recognition was carried out by [3]. These coefficients are highly correlated, and the use of all prediction coefficients may not be necessary for speaker recognition task [6] [7] used a method called orthogonal linear prediction. It is shown that only a small subset of the resulting orthogonal coefficients exhibits significant variation over the duration of an utterance. It is also shown that reflection coefficients are as good as the other feature sets. [8] Used principal spectral components derived from linear prediction coefficients for speaker verification task. Hence a detailed exploration to know the speaker- specific excitation information present in the residual of speech is needed and hence the motivation for the present work.

It has been shown that humans can recognize people by listening to the LP residual signal [9]. This may be attributed to the speaker-specific excitation information present at the segmental (10–30 ms) and suprasegmental levels (1–3 s). The presence of speaker-specific information at the segmental and suprasegmental levels can be established by generating signals that retain specific features at these levels. For instance, speaker-specific suprasegmental information like intonation and duration can be perceived in the signal which has impulses of appropriate strength at each pitch epoch in the voiced region, and at random instances in the unvoiced regions. Instants of significant excitation correspond to pitch epochs in case of voiced speech and some random excitation instants like onset of burst events in case of unvoiced speech. The LP residual has the additional information of the glottal pulse characteristics in the samples between two pitch epochs. Perceptually the signals will be different if these samples (related to the glottal pulse characteristics) are replaced by synthetic model signals [10] [11] [12], [13] or by random noise. It appears that significant speaker specific excitation information is present in the segmental and suprasegmental features of the residual. The present work focuses on extracting speaker-specific excitation information present at the segmental level of the residual.

At the segmental level, each short segment of the LP residual can be considered to belong to one of the five broad categories. They are voiced, unvoiced, plosive, silence and mixed excitation. The voiced excitation is the dominant mode of excitation during speech production. Further, if voiced excitation is replaced by random noise excitation, it is difficult to perceive the speaker's identity [13]. In this paper we demonstrate that the speaker specific characteristics are indeed present at the segmental level of the LP residual, and they can be reliably extracted using hidden Markov models.

The rest of the paper is organized as follows: In Section 2 we examine the speaker specific characteristics of the LP residual, and demonstrate the approach to extract the speaker-specific information from the residual signal. Finally we discuss feature extraction using Melceptral coefficients to capture the speaker specific information from the residual. Section 3 describes parametric approaches such as GMM and HMM based implementation for speaker recognition. Section 4 describes the database used in the study and the performance of speaker recognition systems based on the speaker specific features from the LP residual. The proposed speaker recognition system, based on the LP residual, may not require large amounts of data. Summary and conclusions of this study is presented at the end of the paper.

## 2. SPEAKER CHARACTERISTICS IN THE LP RESIDUAL

Speech signals, as any other real world signals, are produced by exciting a system with source. A simple block diagram representation of the speech production mechanism is shown in the Figure 1. Vibrations of the vocal folds, powered by air coming from the lungs during exhalation, are the sound source for speech. As shown in the Figure 1, the glottal excitation forms the source, and the vocal tract forms the system. The philosophy of linear prediction is intimately related to the basic speech production model. The Linear Predictive Coding (LPC) analysis performs spectral analysis on short segments of speech with an all-pole modelling constraint [14]. Since speech can be modelled as the output of linear, time-varying system excited by a source, LPC analysis captures the vocal tract system information in terms of coefficients of the filter representing the vocal tract mechanism. Hence, analysis of speech signal by linear prediction results in two components, namely the synthesis filter on one hand and the residual on the other hand. In brief, the LP residual signal is generated as a by product of the LPC analysis, and the computation of the residual signal is given below.

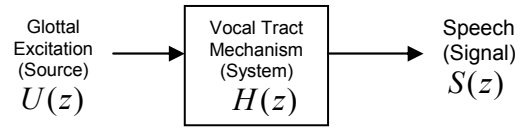

**Figure 1.** Source and System Representation of Speech Production Mechanism.

If the input signal is represented by $u(n)$ and the output signal by $s(n)$, then the transfer function of the system can be expressed as,

$$H(z) = S(z)/U(z) \qquad (1)$$

Where $S(z)$ and $U(z)$ are z-transforms of $s(n)$ and $u(n)$ respectively.

Consider the case where we have the output signal and the system and have to compute the input signal. The above equation can be expressed as

$$S(z) = H(z)U(z) \qquad (2)$$

$$U(z) = S(z)/H(z) \qquad (3)$$

$$U(z) = 1/H(z) \, S(z) \qquad (4)$$

$$U(z) = A(z)S(z) \qquad (5)$$

Where A (z) = 1/ H(z) is the inverse filter representation of the vocal tract system.

### 2.1 Computing LP Residual from Speech Signal

Linear prediction models the output $s(n)$ as the linear function of past outputs and present and past inputs. Since prediction is done by a linear function, the name linear prediction. Assuming an all-pole model for the vocal tract, the signal $s(n)$ can be expressed as a linear combination of past values and some input $u(n)$ as shown below.

$$s(n) = -\sum_{k=1}^{p} s(n-k) + Gu(n) \qquad (6)$$

Where $G$ is a gain factor

Now assuming that the input $U(n)$ is unknown, the signal $S(n)$ can be predicted only approximately from a linear weighted sum of past samples. Let this approximation of $S(n)$ be $\hat{S}(n)$, where

$$\hat{s}(n) = -\sum_{k=1}^{p} a_k s(n-k) \qquad (7)$$

Then the error between the actual value $s(n)$ and the predicted value $\hat{s}(n)$ is given by

$$e(n) = s(n) - \hat{s}(n) = Gu(n) \qquad (8)$$

This error $e(n)$ is nothing but the LP residual of the signal shown in Figure 2.

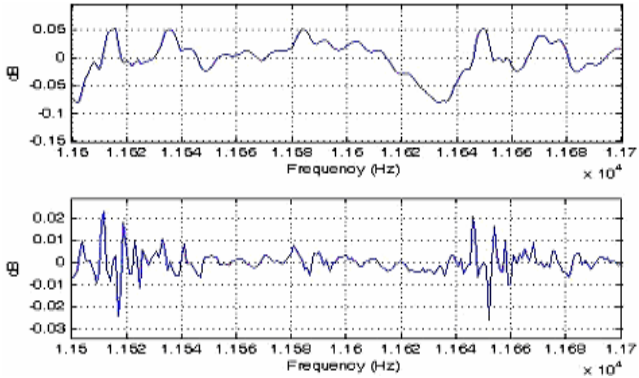

**Figure 2.** Actual signal and its LP residual

### 2.2 Feature extraction of LP residual signal

MFCC features have been used for extracting features from the source signal. MFCC is based on the known variation of the human ear's critical bandwidths with frequency. The MFCC technique makes use of two types of filters namely, linearly spaced filters and logarithmically spaced filters. To capture the phonetically important characteristics of speech, signal is expressed in the Mel frequency scale. This scale has a linear frequency spacing below 1000 Hz and a logarithmic spacing above 1000 Hz. Normal speech waveform may vary from time to time depending on the physical condition of speaker's vocal cords. MFFCs are less susceptible to the said variations [15].

### 2.3 Motivation to use Mel Frequency Cepstral Coefficients (MFCCs):

Since our interest is in capturing global features which correspond to glottal excitation, the low frequency components are to be emphasized. To fulfil this requirement it is felt that MFCC is most suitable as it emphasize low frequency and de-emphasize high frequencies.

### 3. PARAMETRIC APPROACHES

Parametric approaches are model-based approaches. The parameters of the model are estimated using the training feature vectors. It is assumed that the model is adequate to represent the distribution. The most widely used parametric approaches are GMM and HMM based approaches.

### 3.1 Gaussian Mixture Models

GMM is a classic parametric method best used to model speaker identities due to the fact that Gaussian components have the capability of representing some general speaker dependent spectral shapes. Gaussian classifier has been successfully employed in the several text-independent speaker identification applications since the approach used by this classifier is similar to that used by the long term average of spectral features for representing a speaker's average vocal tract shape [16].

As shown in Figure 3 in a GMM model, the probability distribution of the observed data takes the form given by the following equation [17].

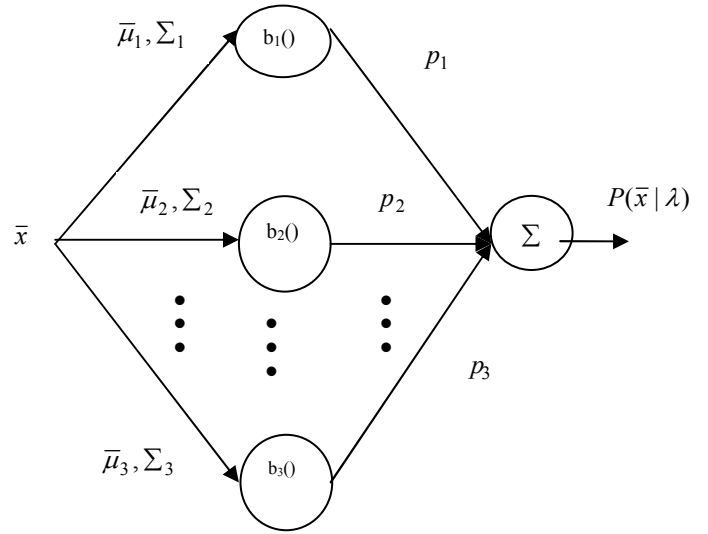

**Figure 3.** Gaussian Mixture Model

$$p(\bar{x} | \lambda) = \sum_{i=1}^{M} p_i b_i(\bar{x})$$

Where M is the number of component densities $\bar{x}$ is a D dimensional observed data (random vector), $b_i(\bar{x})$ are the component densities and Pi are the mixture weights for i = 1... M.

$$b_i(\bar{x}) = \frac{1}{(2\pi)^{D/2} |\Sigma_i|^{1/2}} \exp\left\{-\frac{1}{2}(\bar{x} - \bar{\mu}_i)^T \Sigma_i^{-1} (\bar{x} - \bar{\mu}_i)\right\}$$

Each component density $b_i(\bar{x})$ denotes a D-dimensional normal distribution with mean vector $\bar{\mu}_i$ and covariance matrix $\Sigma_i$. The mixture weights satisfy the condition $\sum_{i=1}^{M} p_i = 1$ and therefore represent positive scalar values. These parameters can be collectively represented as $\lambda = \{p_i, \bar{\mu}_i, \Sigma_i\}$ for i = 1 … M. Each speaker in a speaker identification system can be represented by a GMM and is referred to by the speaker's respective models $\lambda$.

The parameters of a GMM model can be estimated using Maximum Likelihood (ML) [19] estimation. The main objective of the ML estimation is to derive the optimum model parameters that can maximize the likelihood of GMM. Unfortunately direct maximization using ML estimation is not possible and therefore a special case of ML estimation known as Expectation-Maximization (EM) [19] algorithm is used to extract the model parameters. The GMM likelihood of a sequence of T training vectors $X = \{\bar{x}_1,...\bar{x}_T\}$ can be given as [17]

$$p(X|\lambda) = \prod_{t=1}^{T} p(\bar{x}_t|\lambda).$$

The EM algorithm begins with an initial model $\lambda$ and tends to estimate a new model $\bar{\lambda}$ such that $p(X|\bar{\lambda}) \geq p(X|\lambda)$ [17]. This is an iterative process where the new model is considered to be an initial model in the next iteration and the entire process is repeated until a certain convergence threshold is obtained.

### 3.2 Continuous Ergodic Hidden Markov Model for Speaker Recognition

The HMM is a doubly embedded stochastic process where the underlying stochastic process is not directly observable. HMMs have the capability of effectively modelling statistical variations in spectral features. In a variety of ways, HMMs can be used as probabilistic speaker models for both text-dependent and text-independent speaker recognition [20, 21]. HMM not only models the underlying speech patterns but also the temporal sequencing among the sounds. This temporal modelling is advantageous for text-dependent speaker recognition system. Left Right HMM can model temporal sequence of patterns only, where as to capture the patterns of different type ergodic HMM is used [22]

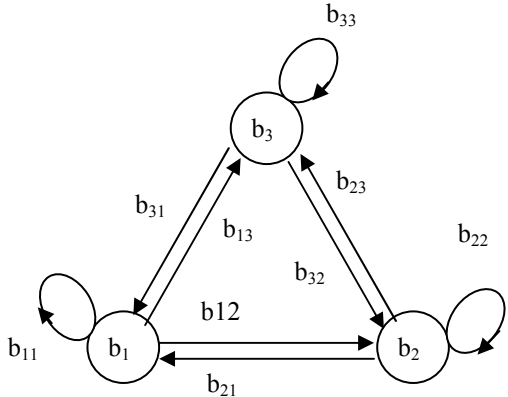

**Figure 4.** Three-State Ergodic HMM.

As shown in the Figure 4 in the training phase, one HMM for each speaker is obtained (i.e., parameters of model are estimated) using training feature vectors. The parameters of HMM are [8] State-transition probability distribution: It is represented by $A = [a_{ij}]$

Where

$$a_{ij} = P(q_{t+1} = j | q_t = i) \quad 1 \leq i,j \leq N \quad (9)$$

The above equation defines the probability of transition from state $i$ to $j$ at time $t$.

For a three state left-right model the state transition matrix is given as 
$$A = \{a_{ij}\} = \begin{bmatrix} a_{11} & a_{12} & a_{13} \\ 0 & a_{22} & a_{23} \\ 0 & 0 & a_{33} \end{bmatrix} \quad (10)$$

The state transition matrix of three state ergodic model is given by

$$A = \{a_{ij}\} = \begin{bmatrix} a_{11} & a_{12} & a_{13} \\ a_{21} & a_{22} & a_{23} \\ a_{31} & a_{32} & a_{33} \end{bmatrix} \quad (11)$$

Observation symbol probability distribution: It is given by $B = [b_j(k)]$ in which

$$b_j(k) = P(O_t = V_k | q_t = j) \quad 1 \leq k \leq M \quad (12)$$

The above equation defines the symbol distribution in state $j = 1,2,3.....N$. The initial state distribution is given by $\pi = P(q_1 = i)$ where

$$\pi_i = P(q_1 = i) \quad 1 \leq i \leq N \quad (13)$$

Here, $N$ is the total number of states, and $q_t$ is the state at time $t$, $M$ is the number of distinct observation symbols per state, and $O_t$ is the observation symbol at time $t$. In testing phase, $P(O/\lambda)$ for each model is calculated, where $O = (O_1 O_2 O_3....O_T)$. Here the goal is to find out the probability for a given model to which the test utterance belongs to. The speaker whose model gives the highest score is declared as the identified speaker. GMM corresponds to a single-state continuous ergodic HMM.

The model parameters can be collectively represented as $\lambda = (A_i, B_i, \pi_i)$ for $i = 1........M$. Each speaker in a speaker identification system can be represented by a HMM and is referred to by the speaker's respective models $\lambda$.

In the testing phase, p (O/λ) for each model is calculated [19]. Where O= (o1o2o3…OT) is the sequence of the test feature vectors. The goal is to find the probability, given the model that the test utterance belongs to that particular model. The speaker model that gives the highest score is declared as the indent.

## 4. EXPERIMENTAL EVALUATION

### 4.1 Database Used for the Study

In general, speaker recognition refers to both speaker identification and speaker verification. Speaker identification is the task of identifying a given speaker from a set of speakers. In the closed-set speaker identification no speaker outside the given set is used for testing. Speaker verification is the task of verifying the identity claim of a given speaker. The result of speaker verification is either to accept or reject the claim of the speaker. In this paper we consider identification task for TIMIT Speaker database.

The TIMIT corpus of read speech has been designed to provide speaker data for the acquisition of acoustic-phonetic knowledge and for the development and evaluation of automatic speaker recognition systems. TIMIT contains a total of 6300 sentences, 10 sentences spoken by each of 630 speakers from 8 major dialect regions of the United States. We consider 200 speakers out of 630 speakers for speaker recognition. Maximum of 30 seconds of speech data is used for training and minimum of 1 second of data for testing. In all the cases the speech signal was sampled at 16 kHz sampling frequency. Throughout this study, closed set identification experiments are done to demonstrate the feasibility of capturing the speaker-specific information from the source features. Requirement of number of mixtures duration of test data to get better accuracy is demonstrated.

### 4.2 Experimental Setup

The system has been implemented in Matlab7 on Windows XP platform. We have used LP order of 12 for all experiments. We have trained the models GMM and HMM using total Gaussian components as 4, 8, 16 and 32 for any training, speech duration of 30 seconds testing is performed using different test speech durations such as 1 second, 3 seconds, and 5 seconds. The same setup has been implemented for both GMM and Ergodic HMM. Here, recognition rate is defined as the ratio of the number of speakers identified to the total number of speakers tested.

### 5. PERFORMANCE EVALAUATION

There is no theoretical way to evaluate the performance of the statistical approaches. To evaluate the speaker recognition system the experiment is carried out for a GMM and 2-state HMM for varying number of Gaussian components such as 4, 8, 16 and 32. Here the model is trained with 30 seconds of speech duration, LP order of 12 and tested with 1 second, 3 seconds and 5 seconds as shown in the Figure 5, 6 and 7 respectively, the ergodic HMM for speaker recognition system outperformed GMM. The experimental results are tabulated in Table 1. The percentage recognition of 2-state ergodic HMM for different Gaussian components such as 4, 8, 16 and 32 seems to uniformly increasing. The minimum number of Gaussian components to achieve good recognition performance seems to be 16 and thereafter the recognition performance is minimal. The recognition performance of the Ergodic HMM drastically increases for the test speech duration of 1 second to 3 seconds Increasing the test speech duration from 3 seconds to 5 seconds improves the recognition performance with small improvement.

### 6. CONCLUSION

In this work we have demonstrated the importance of information in the excitation component of speech for speaker recognition task. Linear prediction residual was used to represent the excitation information. Performance of the recognition experiments shows that 2-state Ergodic Hidden Markov Model can capture speaker-specific excitation information e from the LP residual effectively than GMM. Performance of the system for different HMM states shows that it could capture the speaker-specific excitation information effectively.

The objective in this paper was mainly to demonstrate the capture the speaker-specific excitation information present in the linear prediction residual for speaker recognition effectively than GMM. We have not made any attempt to optimize the parameters of the model used for feature extraction, and also the decision making stage. Therefore the performance of speaker recognition may be improved by optimizing the various design parameters.

**REFERENCES**

[1]   K.N. Stevens, Acoustic Phonetics. Cambridge, England: The MIT Press, 1999.

[2]   O'Shaughnessy, D., 1987. Speech Communication: Human and Machine. Addison-Wesley, New York

[3]   Atal, B.S., 1976. Automatic recognition of speakers from their voices. Proc. IEEE 64 (4), 460–475.

[4]   Makhoul, J., 1975. Linear prediction: a tutorial review. Proc. IEEE 63, 561–580.

[5]   A.E. Rosenberg and M. Sambur, ―New techniques for automatic speaker verification. vol. 23, no.2, pp.169-175, 1975..

[6]   M. R. Sambur, ―Speaker recognition using orthogonal linear prediction, IEEE Trans. Acoust. Speech, Signal Processing, vol. 24, pp.283-289, Aug. 1976.

[7]   J. Naik and G. R. Doddington, ― high performance speaker verification using principal spectral components□, in proc. IEEE Int. Conf. Acoust. Speech, Signal Processing, pp. 881-884, 1986.

[8]   Feustel, T.C., Velius, G. A., Logan, R. J., Human and machine performance on speaker identity verification. Speech Technology, 169-170, 1989.

[9]   Rosenberg, A.E., 1971. Effect of glottal pulse shape on the quality of natural vowels. J. Acoust. Soc. Amer. 49, 583–590.senberg, A.E., 1976. Automatic speaker verification: a review. Proc. IEEE 64 (4), 475–487.

[10]  Ananthapadmanabha G., T.V., Yegnanarayana, B., 1979. Epoch extraction from linear prediction residual for identification of closed glottis interval. IEEE Trans. Acoust. Speech Signal Process. ASSP- 27, 309–319

[11]  Yegnanarayana, B., 1999. Artificial Neural Networks. Prentice-Hall, New Delhi, India.

[12]  Murthy, K.S.R., Prasanna, S.R.M., Yegnanarayana, B., 2004. Speaker-specific information from residual phase. In: Inter nat. Conf

[13]  Furui, S., 1997. Recent advances in speaker recognition. Pattern Recognition Lett. 18, 859–872

[14]  L. R. Rabiner and B. H. Juang, Fundamentals of Speech Recognition. Prentice-Hall, 1993.

[15]  Gish, H., Krasner, M., Russell, W., and Wolf, J., "Methods and experiments for text-independent speaker recognition over telephone channels," Proceedings of the IEEE International Conference on Acoustics, Speech, and Signal Processing (ICASSP), vol. 11, pp. 865-868, Apr. 1986.

[16]  Reynolds, D. A., and Rose, R. C., " Robust Text-Independent Speaker Identification using Gaussian Mixture Models,'' IEEE-Transactions on Speech and Audio Processing, vol. 3, no. 1, pp. 72-83,1995.

[17]  A. P. Dempster, N. M. Laird, and D. B. Rubin, "Maximum likelihood from incomplete data via the EM algorithm", J. Royal Statist. Soc. Ser. B. (methodological), vol. 39, pp. 1-38, 1977

[18]  M. Forsyth and M. Jack, ―Discriminating semi-continuous HMM for speaker verification,☐ in proc. IEEE Int. Conf. Acoust. Speech, Signal Processing, vol.1, pp. 313-316, 1994.

[19]  M. Forsyth, ―Discriminating observation probability (DOP) HMM for speaker verification, Speech Communicaiton, vol. 17, pp.117-129, 1995.

[20]  R. Rajeshwara Rao, "Automatic Text-Independent Speaker Recognition using source feature", Ph.D Thesis (submitted, Jan-2010).

**BIOGRAPHY**

**R.Rajeswara Rao** received his B.Tech from Nagarjuna University and M.Tech from JNT University degrees in Computer Science and in 1999 and 2003, respectively. He is currently pursuing Ph.D. from JNT University, Hyderabad, India since 2004. His research areas of interest are Speech Processing, Neural Networks, and Pattern Recognition.

**V. Kamakshi Prasad** received his M.Tech from Andhra University and Ph.D. from IIT-M. He is having 16 years of teaching experience. His research interest include Speech Processing, Image Processing, Neural Networks, and Pattern Recognition. He has published 40 research papers in national and international journals.

**A. Nagesh** has received B.Tech and M.Tech from Osmania University, in computer science and engineering in 1996 and 2002, respectively. He is currently pursing Ph.D. from JNT University, Hyderabad. Since 2004 his research areas of interest are speech processing and pattern recognition.

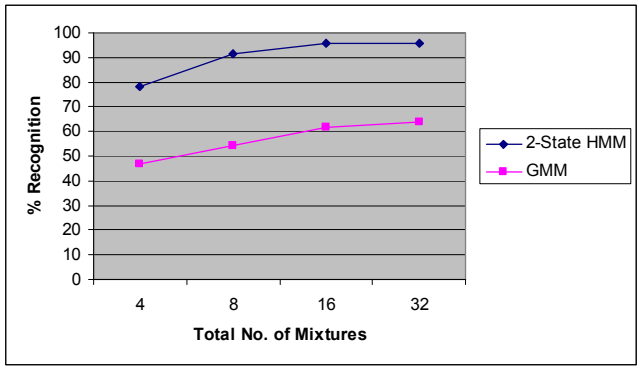

**Figure 5.** Speaker Recognition Performance for Test Speech duration of 1Second.

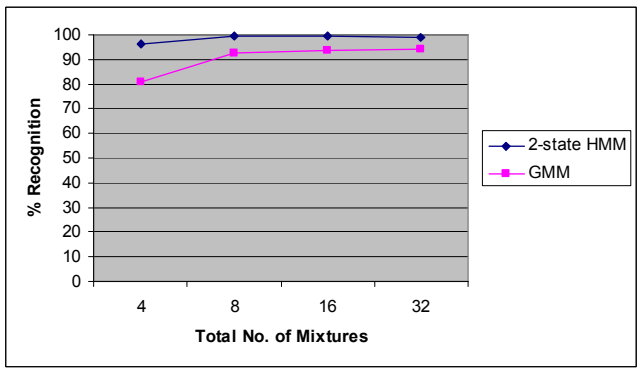

**Figure 6.** Speaker Recognition Performance for Test Speech duration of 3 Seconds.

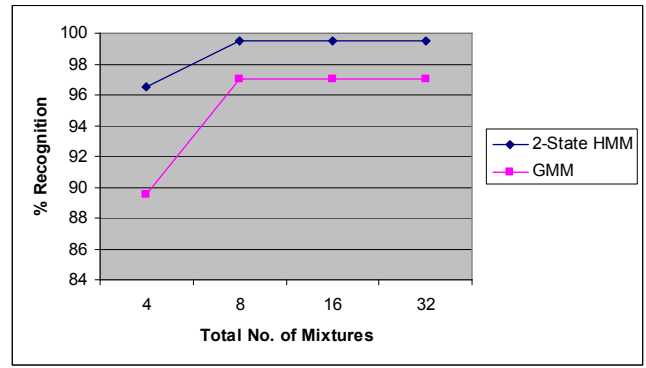

**Figure 7.** Speaker Recognition Performance for Test Speech duration of 5 Seconds.

**Table 1.** Speaker Recognition Performance for two statistical approaches

| No. of Mixture Components | Recognition rate (%) | | | | | |
|---|---|---|---|---|---|---|
| | Testing Speech duration | | | | | |
| | 1 Sec. | | 3 Sec. | | 5 Sec. | |
| | GMM | HMM | GMM | HMM | GMM | HMM |
| 4 | 47 | 78 | 81 | 96.5 | 89.5 | 96.5 |
| 8 | 54 | 91.5 | 92.5 | 99.5 | 97 | 99.5 |
| 16 | 61.5 | 95.5 | 93.5 | 99.5 | 97 | 99.5 |
| 32 | 64 | 96 | 94 | 99 | 97 | 99.5 |